\documentclass[letterpaper, 10 pt, conference]{ieeeconf}  

\IEEEoverridecommandlockouts                             
\usepackage{graphics}
\usepackage{epsfig} % for postscript graphics files
\usepackage{mathptmx} % assumes new font selection scheme installed
\usepackage{times} % assumes new font selection scheme installed
\usepackage{amsmath} % assumes amsmath package installed
\usepackage{amssymb}  % assumes amsmath package installed
\usepackage{subcaption}
\usepackage{graphicx}
\usepackage{float}
\usepackage{caption}
\usepackage{mathtools}
\usepackage{stfloats}
\usepackage{mathrsfs}
\usepackage{multirow}
\usepackage{makecell}
\usepackage{vcell}
\usepackage{booktabs}
\usepackage{diagbox}
\usepackage{csquotes}
\usepackage{cite}

\usepackage[hyphens]{url}
\usepackage[hidelinks]{hyperref}
\hypersetup{breaklinks=true}
\usepackage{cite}

\DeclareMathOperator{\arctantwo}{arctan2}

\usepackage{hyperref}
\hypersetup{
colorlinks=true,
linkcolor=blue,
filecolor=magenta,
urlcolor=black,
}

\title{\LARGE \bf
Autonomous Drone Delivery to Your Door and Yard
}
\author{Shyam Sundar Kannan and Byung-Cheol Min% <-this % stops a space
\thanks{The authors are with SMART Lab, Department of Computer and Information Technology, Purdue University, West Lafayette, IN 47907, USA \tt\small{kannan9@purdue.edu | minb@purdue.edu}}%
}
\begin{document}
\maketitle
\thispagestyle{empty}
\pagestyle{empty}

%%%%%%%%%%%%%%%%%%%%%%%%%%%%%%%%%%%%%%%%%%%%%%%%%%%%%%%%%%%%%%%%%%%%%%%%%%%%%%%%
\begin{abstract}
In this work, we present a system that enables delivery drones to autonomously navigate and deliver packages at various locations around a house according to the desire of the recipient and without the need for any external markers as currently used. This development is motivated by recent advancements in deep learning that can potentially supplant the specialized markers presently used by delivery drones for identifying sites at which to deliver packages. The proposed system is more natural in that it takes instruction on where to deliver the package as input, similar to the instructions provided to human couriers. First, we propose a semantic image segmentation-based descending location estimator that enables the drone to find a safe spot around the house at which it can descend from higher altitudes. Following this, we propose a strategy for visually routing the drone from the descent location to a specific site at which it is to deliver the package, such as the front door. We extensively evaluate this approach in a simulated environment and demonstrate that with our system, a delivery drone can deliver a package to the front door and also to other specified locations around a house. Relative to a frontier exploration-based strategy, drones using the proposed system found and reached the front doors of the 20 test houses $161\%$ faster.
\end{abstract}

%%%%%%%%%%%%%%%%%%%%%%%%%%%%%%%%%%%%%%%%%%%%%%%%%%%%%%%%%%%%%%%%%%%%%%%%%%%%%%%%
\section{Introduction}
\label{sec:intro}
In recent years, the use of drones for the delivery of packages and food has gained significant attention, primarily due to drones' operational efficiency and their ability to reach remote areas. Consequently, delivery drones have become prevalent amongst logistics undertakings, not only allowing for reduced costs but also speeding the delivery process \cite{floreano2015science}. Delivery of packages using drones involves various challenges such as the design of the robot, navigation, safety, and ultimately delivering the package at a location desirable to the recipient. Previous studies have primarily focused on the aspects of design \cite{kornatowski2017origami, kornatowski2020morphing}, navigation \cite{munoz2019deep, kim2018cbdn, brunner2019urban} and the safety \cite{schenkelberg2016reliable, altawy2016security}; there has been less investigation on delivering packages at specific, recipient-chosen locations around a house, despite this being a critical part of automating delivery by drones. 

\begin{figure}[t] 
    \centering
    \includegraphics[width=0.98\columnwidth]{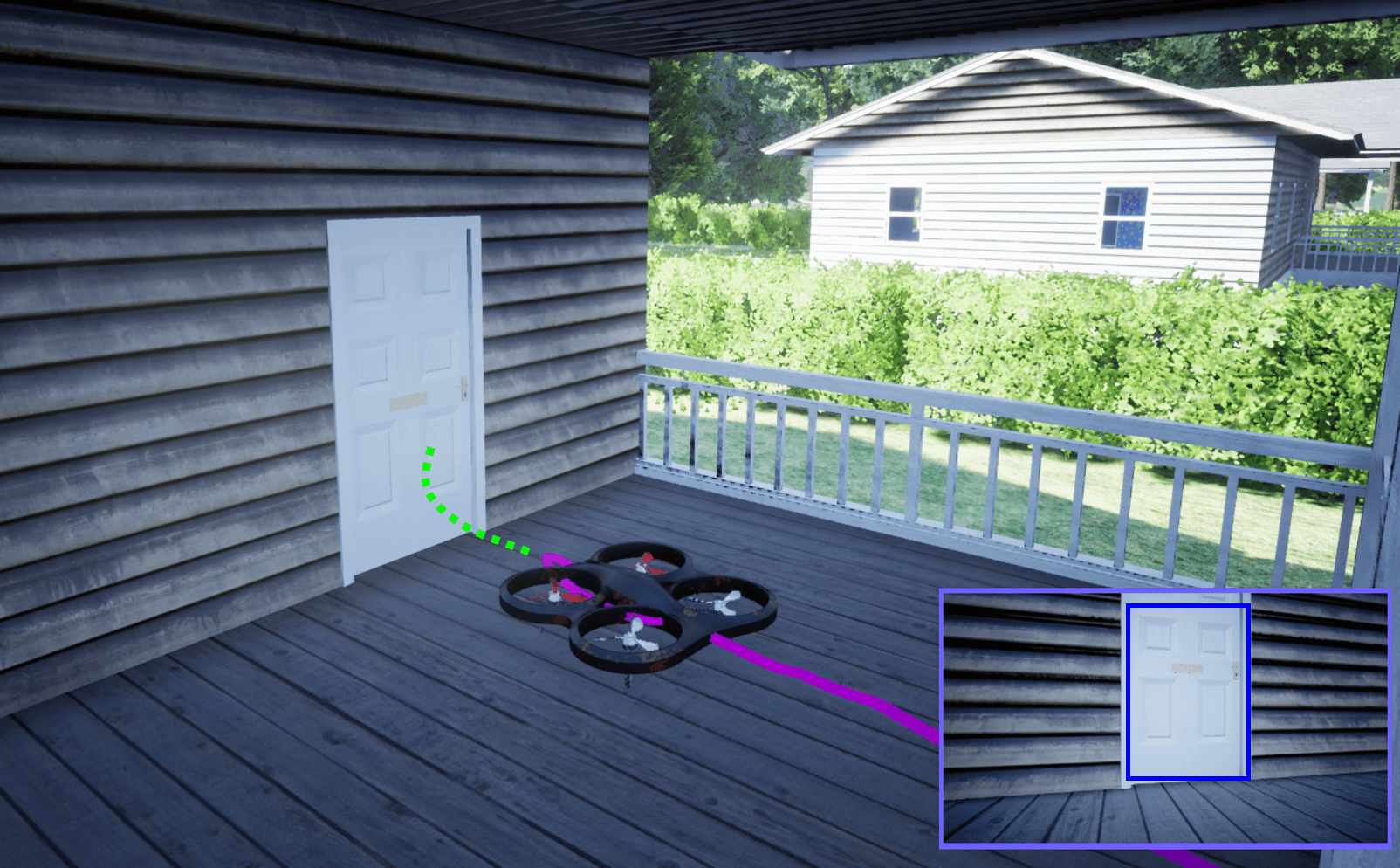}
    \caption{A drone routing itself towards the front door of a house to deliver a package. The trajectory traced by the drone marked in purple and the path to be followed in green. The front view of the drone along with the door detected (blue box) is shown at the right bottom.} 
    \label{fig:simple_house_expt}
    % \vspace{-5mm}
\end{figure}

Many e-commerce retailers and fast-food chains have been exploring the utility of drones for delivery of their products. Most of these drones fly to a destination and then drop the package far away from the doorstep, necessitating that the human recipient collects the package immediately upon its delivery. Delivery drones from Amazon \cite{Amazonco52:online}, UPS \cite{UPSDrone21:online}, Domino's \cite{pepitone2013domino}, and Walmart \cite{WalmartN82:online} all use a similar approach where packages are placed on a specific landing pad or lowered and dropped using a string. These methods consistently require human intervention in receiving the package. However, in most real-world scenarios, a human courier delivers a package at a doorstep while the recipient is not available to receive it. In short, the need for recipient presence during drone delivery can potentially limit the large-scale deployment of this technology.

At present, the literature contains several interesting studies on the impact of delivery drones in terms of ethics and privacy\cite{luppicini2016technoethical}, mobility strategies \cite{yoo2018}, economic impacts \cite{rao2016societal}, and the threats they pose \cite{shelley2016model}. In contrast, only very limited field studies have been conducted regarding the practical aspects of delivery by drones. One extant paper focused on delivering packages to the balconies of apartment houses \cite{brunner2019urban}, for which the delivery drone used visual markers to detect the balcony and also to localize. However, that approach is strictly limited to apartments with balconies. Many types of residential housing exist, including not only apartments but also detached houses and townhouses; moreover, the majority of the United States population resides in detached single-unit houses that do not usually have a balcony \cite{Themostp53:online}. Methods utilizing reinforcement learning have been used for routing the delivery drone to the recipient's house while avoiding obstacles \cite{munoz2019deep}. But, the drone delivers the package in the proximity of the house and not at a specific location. A recent user study found that recipients of packages prefer their deliveries to be placed close to the door rather than far away from it \cite{kannan2021}. Hence, in order to enable large-scale deployment of package delivery using drones, robots should be able to deliver packages to a doorstep or any other preferred location, just like a human courier.

A field study on user acceptance of delivery drones in a campus environment found that people overall have a positive attitude towards delivery drones \cite{kornatowski2018last}. However, human interventions were involved in the safe landing of the delivery drones that may not be well suited for complete autonomous delivery. Methods for the autonomous safe landing of unmanned aerial vehicles (UAVs) have been proposed, such as using neural networks capable of identifying flat regions to estimate safe landing spots for UAVs \cite{marcu2018safeuav, hinzmann2018free}. Other approaches for the identification of safe landing spots include fuzzy logic \cite{patterson2014timely}, naive Bayes classifiers\cite{li2013software}, and optical flow analysis \cite{cesetti2010vision}. Despite progress in this area, these methods are oriented towards safe landing during emergencies; they are not applicable for safe landing in proximity to a house during routine delivery tasks. 
 
In this paper, we present an autonomous system for the last-mile delivery of packages using drones. Distinct from previous works, the proposed system delivers the package to a specific location around the house as desired by the recipient. For instance, some people might prefer a drone to deliver packages at the doorstep like a human courier, while others might prefer delivery to their back yard, and so on. Our approach enables a drone not only to deliver to the desired recipient-indicated site but also to perform that task without the need for any external markers like a landing pad or a fiducial marker. First, the proposed system enables drones to perceive features in an aerial image of the delivery site and use them to estimate a safe descent spot. Then, the drone descends at that location and performs vision-based navigation to find its way to the final package delivery spot. Fig. \ref{fig:simple_house_expt} depicts a drone navigating to the front door of a house, where it has been directed to deliver the package. 
 
The main contributions of this paper are the following:
\begin{itemize}
    \item A novel external-marker-free autonomous delivery drone capable of delivering packages anywhere around a house.   
    \item A semantic image segmentation-based descent location estimation method that enables a drone to identify a safe spot for its descent that is in the vicinity of the recipient's house and close to the final package delivery spot.
    \item  A real-time path planner that can route a drone from its descent location to the final delivery spot.
    \item Extensive experimentation and evaluation of the proposed system in a photo-realistic simulator with respect to the real-world variation a drone might encounter during a delivery task.
    \item Demonstration of drones reaching the front doors of the houses $161\%$ faster than when using an exploration-based method.
\end{itemize}

\section{Problem Description}
\label{sec:problem}
The overall problem considered in this paper is the autonomous delivery of packages by drones to specific locations such as the front door or back yard of a house as designated by the recipient and without the use of additional specialized markers. The core assumption of this work is that the drone delivers the package to a typical single-unit house (the most common house type in the United States) and that the specified delivery location is a feasible one. 

The last-mile delivery problem considered in this work begins once the drone is above the roof of the recipient's house. Navigation to the top of a house can be achieved effortlessly using autopilot systems that allow for point-to-point navigation at higher altitudes where good Global Positioning System (GPS) reception is available \cite{meier2015px4}. The challenge, then, is to empower the drone to navigate from that overhead point to a specific location like the front door, at which it is to deliver the package. We also assume that the drone delivery is made during conditions with good visibility such as in bright daylight and that the visibility is not affected by external factors such as the weather. We formulate this as a vision-based navigation problem, where the drone navigates to the final delivery location based on visual perception. Drone design aspects involved in actually delivering the package at the designated location are not considered within the scope of this work. For simplicity, we only consider the following delivery locations: front door, back yard, front yard, and front paved area. 

\begin{figure*}[t]
\centering
\includegraphics[width=1\textwidth]{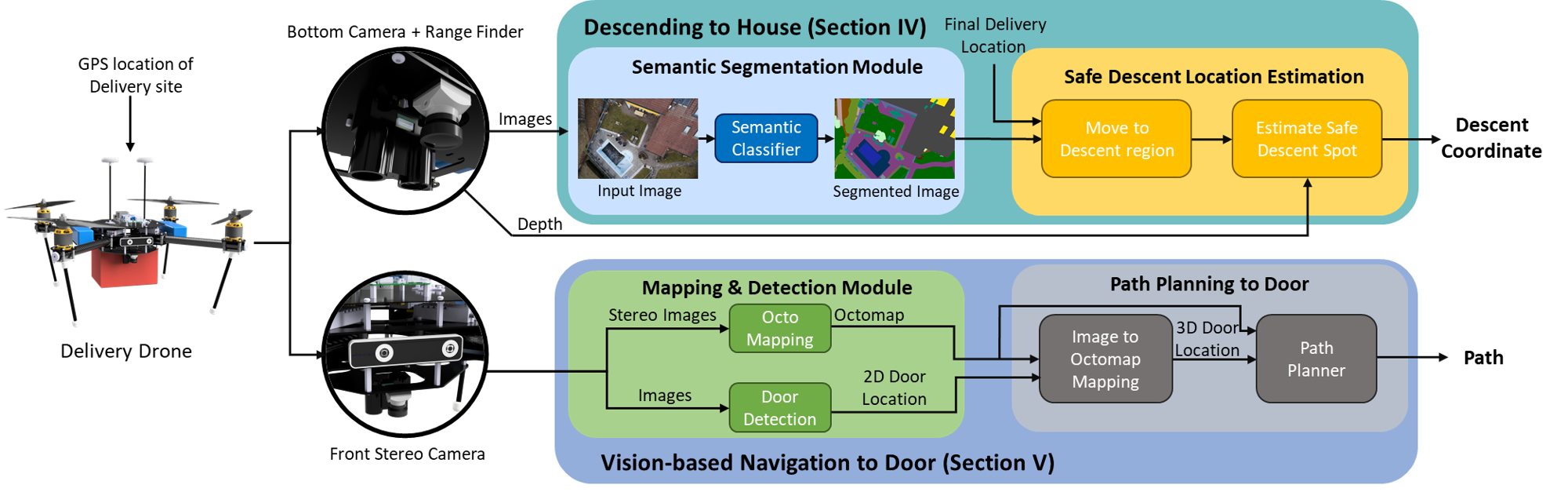}
\caption{Schematic representation of the overall system. The delivery drone is equipped with a bottom camera,  a front-facing stereo camera, and a range finder. The bottom camera is used to estimate the safe descent spot based on the segmented aerial image and the features observed. The front stereo camera is used by the drone to identify the door of the recipient's house and to map the environment that in turn assists the drone while navigating to the door.}
\label{img:overall_system}
%\vspace{-3mm}
\end{figure*} 

\section{System Overview}
\label{sec:sys_overview}
The proposed system enhances the automation of the delivery drones by facilitating them to deliver packages at a specific location similar to human couriers. In other words, the proposed system routes the drone from the top of a house to a very specific delivery spot. The drone can potentially use GPS-based navigation to reach the top of the house since the GPS remains accurate at the higher altitudes \cite{meier2015px4}. The specific location where the drone needs to deliver the package is given as input to the system. This is similar to the instruction given to the logistics firm about where exactly the recipient wants the package to be placed. %In this study, we consider different delivery spots like the front door, backyard, and front yard. 

As shown in Fig. \ref{img:overall_system}, the proposed system consists of two key components: 1) descending to the house through segmented aerial image, and 2) vision-based navigation to the delivery spot. Descending to the house component computes a safe spot for the drone to descend itself from the top of the house. The descent location estimated is also such that the drone can easily reach the final delivery spot from that descent location. For instance, if the drone needs to deliver the package to the front door of a house, it may not be feasible to reach the front door if the drone descends into the back yard of a house. Ideally, the drone should descend in the front yard or at some paved areas in front of the house. Semantic segmentation is performed on the aerial images to identify the features present such as the roof, grass, trees, paved area, and so on. Based on these features observed and the final delivery spot, the drone estimates a feasible location where it can descend itself. In case, the descent location is the delivery location itself (like back yard), then the drone completely lands and delivers the package. The complete details on the implementation of the descending to house system are explained in Section \ref{sec:land}.

The vision-based navigation system assists the drone in autonomously navigating from the descending location to the final delivery spot during scenarios where the descending location and the delivery spot are not the same. The drone first tries to identify the final delivery spot such as the door using visual perception. If the final delivery spot is not evidently visible, the drone performs a search until it finds the delivery location. Then, the drone navigates itself to that location and descends itself to drop the package. The drone while searching or navigating to the final delivery spot maps the environment and also dynamically routes itself. The complete details about the functioning of the navigation system are described in Section \ref{sec:navigation}.

The drone platform considered for this paper is equipped with a camera and a range sensor facing downward. The downward-facing camera is used to obtain an aerial view of the house and estimate the location to descend the drone. The range sensor is used to estimate the height from the ground plane and assists the drone in the descending process. The drone is also equipped with a front-facing stereo camera. The front-facing stereo camera enables the drone to map and navigate in the proximity of the house post descending. The perception sensors for the drone have been selected based on the sensors that are commonly used with the drones for various applications. 

\section{Descending to the House}
\label{sec:land}
In this section, we describe the method used by the delivery drone for identifying a safe descending position based on the segmented aerial images obtained from the drone's downward-facing camera. The various steps involved in computing a safe descending position from the aerial images are shown in Fig. \ref{fig:lowering}.

\subsection{Descent Location Estimation}
\begin{figure*}[t]
\centering
    \begin{subfigure}{0.245\linewidth} % input image
        \includegraphics[width=\linewidth]{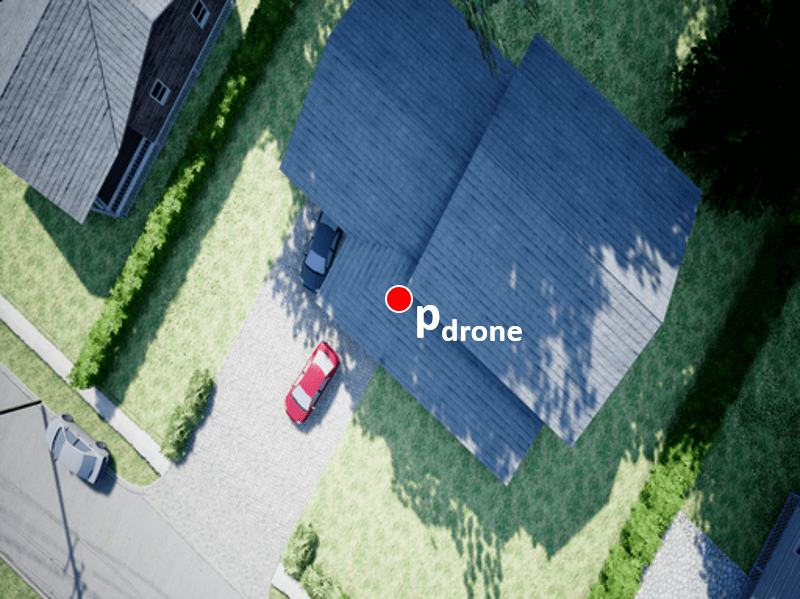}
        \caption{}
        \label{fig:seciv_a}
    \end{subfigure}\hfill
    \begin{subfigure}{0.245\linewidth} % input image
        \includegraphics[width=\linewidth]{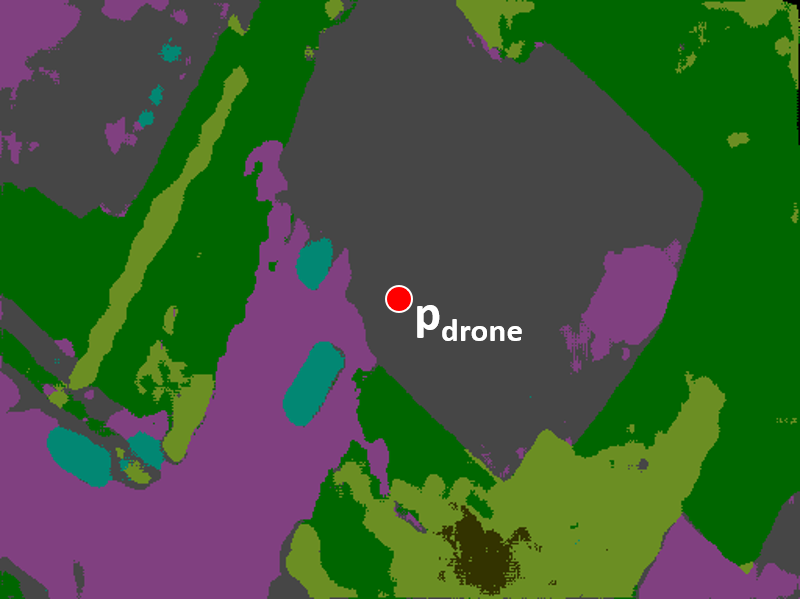}
        \caption{}
        \label{fig:seciv_b}
    \end{subfigure}\hfill
    \begin{subfigure}{0.245\linewidth} % input image
        \includegraphics[width=\linewidth]{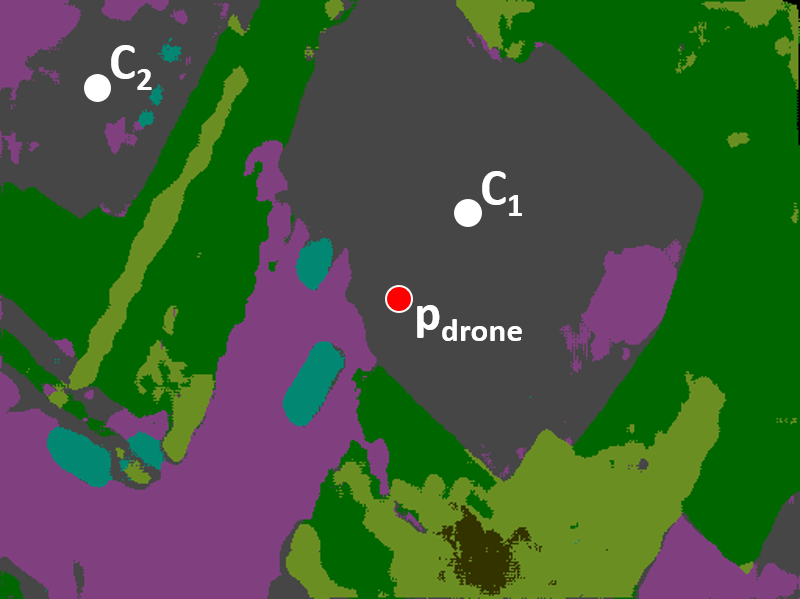}
        \caption{}
        \label{fig:seciv_c}
    \end{subfigure}\hfill
    \begin{subfigure}{0.245\linewidth} % input image
        \includegraphics[width=\linewidth]{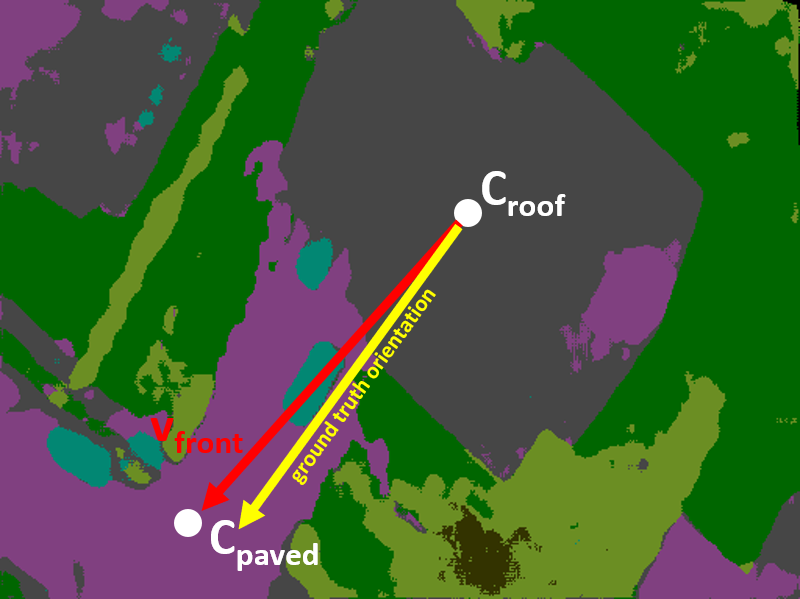}
        \caption{}
        \label{fig:seciv_d}
    \end{subfigure}\hfill
    \begin{subfigure}{0.245\linewidth} % input point cloud
        \includegraphics[width=\linewidth]{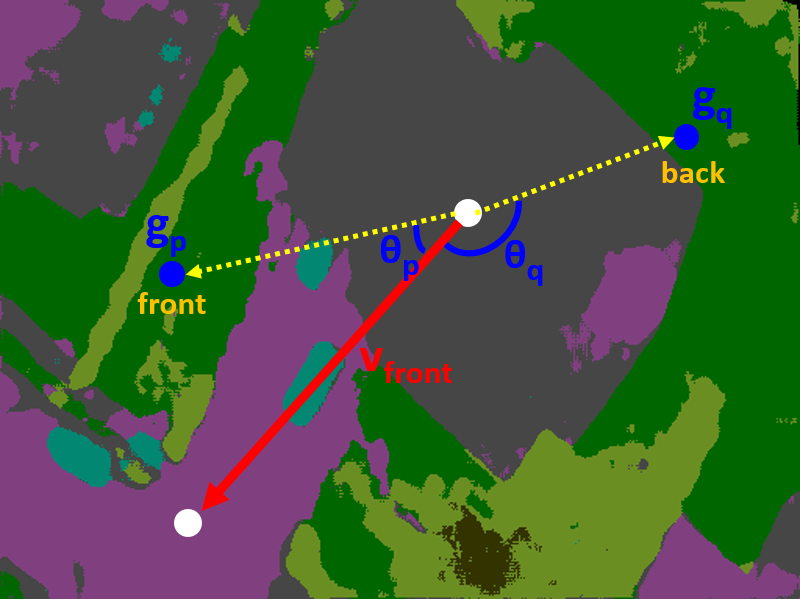}
        \caption{}
        \label{fig:seciv_e}
    \end{subfigure}\hfill
    \begin{subfigure}{0.245\linewidth} % remove noise from the input
        \includegraphics[width=\linewidth]{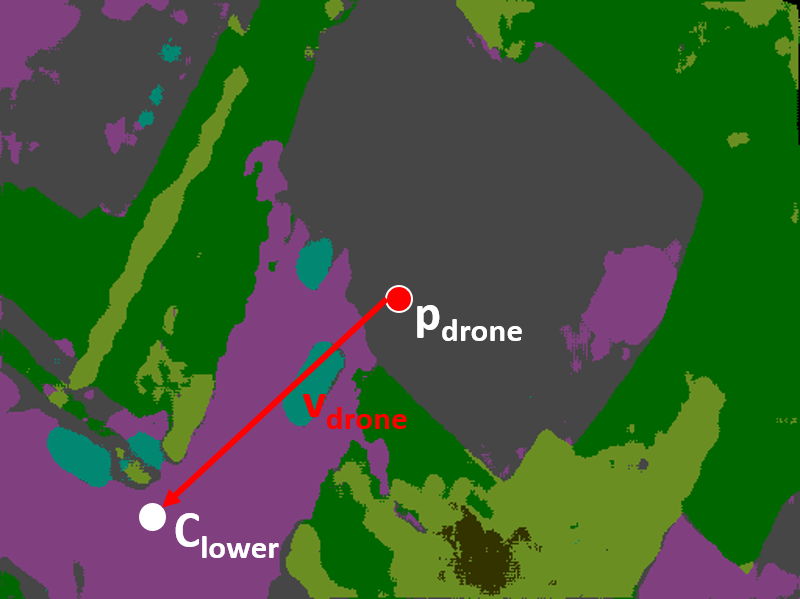}
        \caption{}
        \label{fig:seciv_f}
    \end{subfigure}\hfill
    \begin{subfigure}{0.245\linewidth} % decide target points 
        \includegraphics[width=\linewidth]{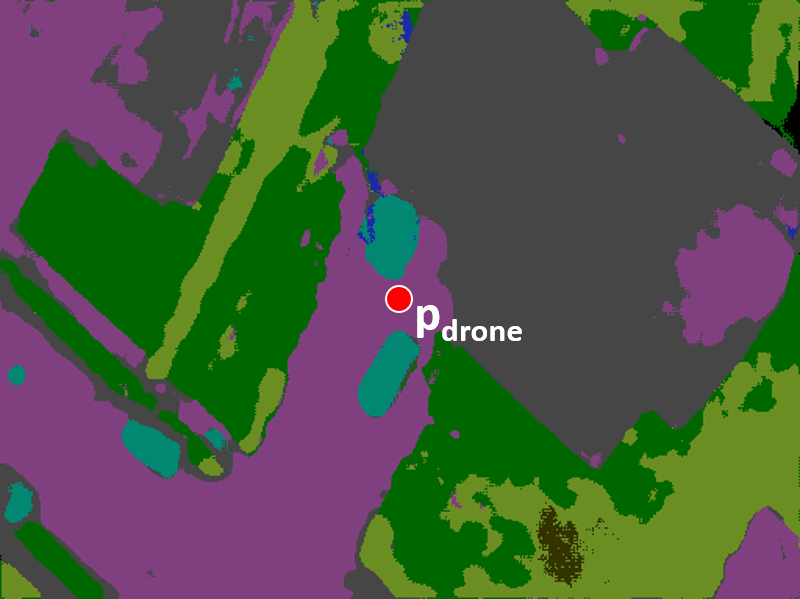}
        \caption{}
        \label{fig:seciv_g}
    \end{subfigure}\hfill
    \begin{subfigure}{0.245\linewidth}% decide target point
        \includegraphics[width=\linewidth]{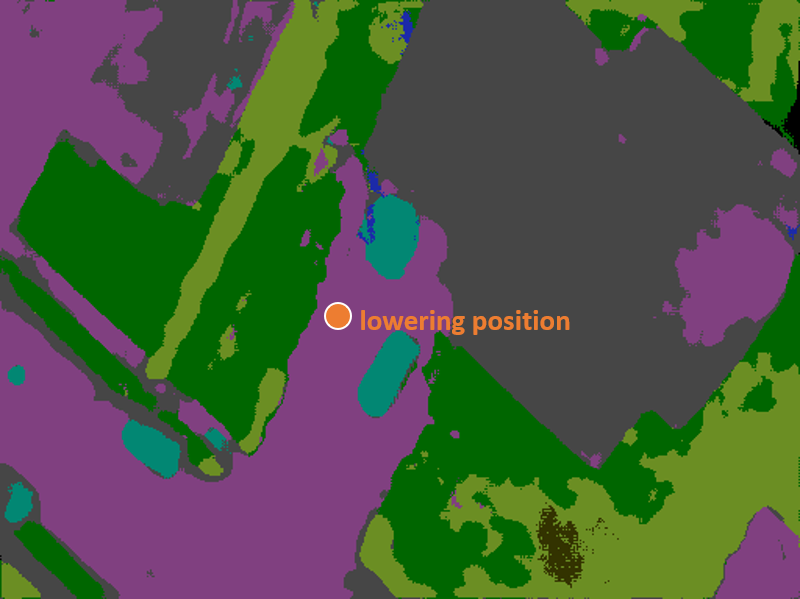}
        \caption{}
        \label{fig:seciv_h}
    \end{subfigure}\hfill
\caption{Various steps involved in estimating a safe descending position from the aerial images: (a) The aerial image of the recipient's house; (b) The segmented results of the aerial image; (c) The drone identifying the roof of the recipients house from the segmented results; (d) Estimated orientation of the recipient's house (red arrow) along with the ground truth orientation (yellow arrow); (e) Classifying regions of grass as front yard and back yard; (f) Drone direction of motion to move towards the descending region; (g) Updated position of the drone after moving to the descending region; and (h) Descent position computed (orange dot).}
\label{fig:lowering}
%\vspace{-7mm}
\end{figure*}

Descent location estimation is performed once the drone reaches the GPS coordinate of the recipient's house. The GPS coordinate to which the drone arrives can be anywhere over the house, and hence, the drone cannot descend itself right away. The drone captures an aerial view of the recipient's house using the downward-facing camera. The camera follows a pinhole model and captures an image of size $W\times H$. In the image frame, the drone is located at the center of the image, and its location is given as $p_{drone} = (W/2, H/2)$. In Fig. \ref{fig:seciv_a} the position of the drone, $p_{drone}$ is marked using a red dot.

% \noindent\textbf{Semantic Segmentation of Aerial Image.~}
Semantic segmentation of the aerial image is used to obtain locations of different regions and also the various objects present around the house in the image frame. From the segmentation, the locations of the roof, paved area\footnote{Note: The semantic classifier considers both roads and pavements as paved area.} and grass are extracted. The information about these regions is later used in estimating the descent location. The location of other objects such as trees, vegetation, car, and fence are also extracted and are collectively considered as obstacles. The details about the location of the obstacles help in identifying the regions to stay away from while descending. Fig. \ref{fig:seciv_b} shows the semantic image segmentation for the aerial image shown in Fig. \ref{fig:seciv_a} based on the color labeling shown in Fig. \ref{fig:sem_seg}.

\noindent\textbf{Roof Identification.~}
First, the drone identifies the roof of the house to which it needs to deliver the package based on the region corresponding to the roof class from the segmented image. If there are multiple houses (recipient's house and neighboring houses) present in the image frame, then there will be multiple segments corresponding to roof class in the segmented image. Let $n_{roof}$ be the total number of roof segments in the segmentation result and $c_i, i \in \{1,...,n_{roof}\}$ be the centroids of those roof segments. The roof segment corresponding to the recipient's house is identified as:
\begin{equation}
    \min_i^{n_{roof}} \|{c_i - p_{drone}}\|.
\end{equation}

In other words, the roof segment with the centroid closest to the drone's position (center of the image) is selected as the roof of the recipient's house. This is valid since the drone is at a GPS coordinate that is close to the proximity of the recipient's house than the neighboring houses. Fig. \ref{fig:seciv_c} shows two roof segments with centroids $c_1$ and $c_2$ marked as white dots. The roof with centroid $c_1$ is closer to $p_{drone}$ and hence, the same is the roof of the recipient's house.

\noindent\textbf{House Orientation Estimation.~} 
The semantic bounds of the recipient's house's roof give the footprint of the house but do not give any information about the orientation of the house. The orientation of the house with respect to the road is fundamental for locating the various features present around the footprint of the house. For instance, if the final delivery spot is the front door, it is important to know the front side of the house. Also, the front and the back yard will be classified as $grass$ by the semantic classifier. The details about the house's orientation are needed to further classify the grass segments as the front and the back yard. 

The orientation of the house is estimated using the bounds of the roof and paved area segments. The fact that houses have paved areas connecting the front of the house and the road is used for estimating the house's orientation. The results of the semantic classification can have multiple segments corresponding to paved areas. The paved area segment that shares a boundary with the house's roof segment and also extends to the boundary of the image is the paved area segment that leads to the front side of the house. The paved area in front of the house leads to the house's footprint and hence they share a boundary with the roof segment. The portions corresponding to the road in the paved area keep extending and it goes on to the bounds of the image. Let $c_{roof}$ and $c_{paved}$ be the centroids of the roof segment of the house and the paved area leading to the house. Now, an approximate direction of the house's front side in the image frame ($\vec v_{front}$) can be estimated as:
\begin{equation}
    \vec v_{front}= c_{roof} - c_{paved}.
\end{equation}

Fig. \ref{fig:seciv_d} shows the direction of the house's estimated orientation in red arrow, along with the actual orientation marked in yellow. 

\noindent\textbf{Front and Back Categorization.~}
In scenarios where the drone needs to deliver the package to the front or the back yard, it is ideal if the drone descends into the corresponding yard. But the regions corresponding to both the yards will be labeled as $grass$ and are not distinguishable from the results of semantic classification. Now, the angle between the pixels in the image labeled as grass and $\vec v_{front}$ is used to classify whether they lie in the front or back of the house. Let $n_{grass}$ be the total number of pixels in the image labeled as grass and $g_i \in \{1,...,n_{grass}\}$ be the coordinate of those pixels. Now, the angle these pixels form with the orientation of the house is computed as:
\begin{equation}
    \theta_{i} = \arctantwo(g_{i_{y}}, g_{i_{x}}) - \arctantwo(v_{{front}_y}, v_{{front}_x})
\end{equation}

\noindent where $v_{{front}_y} = c_{{roof}_y} - c_{{paved}_y}$, and $v_{{front}_x} = c_{{roof}_x} - c_{{paved}_x}$.

Then, the angles are normalized to be in the range $(-\pi,\pi]$ as:
\begin{equation}
    \theta_{i} = 
\begin{cases}
    \theta_{i} - 2\times\pi, & \text{if } \theta_{i} > \pi\\
    \theta_{i} + 2\times\pi, & \text{if } \theta_{i} \leq -\pi\\
    \theta_{i}, & \text{otherwise}.
\end{cases}
\end{equation}

Based on the normalized values of $\theta_{i}$, the pixels labeled as grass are classified as:
\begin{equation}
\begin{aligned}
    \textit{front}, & \text{  if } -\pi/2 \leq \theta_{i} \leq \pi/2 \\
    \textit{back},  & \text{  otherwise}.
\end{aligned}
\end{equation}

Fig. \ref{fig:seciv_e} shows two random pixels $g_{p}$ and $g_{q}$ labelled as grass. Based on the angle these pixels form with $\vec v_{front}$, $g_{p}$ is classified as $front$ and $g_{q}$ as $back$.

\noindent\textbf{Motion to Descent Region.~}
The drone can be flying over any region around the house based on the GPS coordinate of the delivery site provided. The region above which the drone is currently flying can be identified using the semantic class label of the pixel $(p_{drone_x}, p_{drone_y})$. Now, the drone needs to move to the region where it needs to descend. Based on the final delivery spot $\bigstar\in \{\textit{front door, front paved area, back yard, front yard}\}$, the descent region $\in \{\textit{front paved area, back yard, front yard}\}$ is decided as:
\begin{equation*}
\begin{aligned}
    \textit{front paved area}, & \text{  if } \bigstar=\text{front door or front paved area} \\
    \textit{back yard},  & \text{  if } \bigstar=\text{back yard} \\
    \textit{front yard},  & \text{  if } \bigstar=\text{front yard}.
\end{aligned}
\end{equation*}

In other words, the drone descends on the paved area if the final delivery spot is the front paved area or the front door. The drone descends in the corresponding yard if the front or the back yard is specified as the final delivery spot. If the drone is currently not flying over the region where it needs to descend, it moves towards that region. Let $c_{descend}$ be the centroid of the region where it needs to descend, then the direction of motion of the drone ($\vec v_{drone}$) is computed as:
\begin{equation}
     \vec v_{drone} =  p_{drone} - c_{descend}.
\end{equation}
Fig. \ref{fig:seciv_f} shows the direction $\vec v_{drone}$ (red arrow) for the sample environment considered. The drone moves in the direction $\vec v_{drone}$ while maintaining a fixed altitude. The drone keeps moving in that direction until it is over the descent region. The drone uses the new semantic segmented images obtained from the downward-facing camera as it moves and stops the motion once the region below the drone at pixel $(p_{drone_x}, p_{drone_y})$ is labeled as the region where the drone needs to descend. 

\noindent\textbf{Safe Descending.~}
Now, that the drone is above the region where it needs to descend, the drone needs to estimate a safe spot in that region where it can descend itself. Fig. \ref{fig:seciv_g} shows the updated position of the drone where it is flying over the paved area region which is the region where it needs to descend. First, the drone measures its flying height ($h_{drone}$) over that region using the range finder. The height is measured so as to build a back-projection model to find the correspondences between the pixels in the image frame and its actual 3D coordinate in the camera frame. A pixel $(u,v)$ in the image frame can be mapped to the corresponding location in the camera frame $(U, V, W)$ as:
\begin{equation}
\begin{aligned}
    \textit{U} & = h_{drone} \times (\text{u} - O_{x}) / f_{x}\\
    \textit{V} & = h_{drone} \times (\text{v} - O_{y}) / f_{y}\\
    \textit{W} & = h_{drone}
\end{aligned}
\label{eqn:mapping}
\end{equation}
where $(O_{x}, O_{y})$ is the optical center of the camera, and $( f_{x}, f_{y})$ is the focal length of the camera along $x$ and $y$ axes.

Safe descent location is selected such that the descending coordinate is close to the footprint of the house and also at a safe distance from the obstacles. Prior research has found that the drone can drift up to $2~m$ while descending due to errors in the GPS reception and estimation \cite{kornatowski2018last}. Hence, in the implementation, it was set such that the descending location is at least $2.5~m$ from the house's footprint and the other obstacles such as cars, vegetation, and so on. 

Using the image frame to camera frame mapping as shown in Eq. \ref{eqn:mapping}, the distance to the closest point on the roof and other obstacles from the drone is measured. Let $d_{roof}$ be the distance between the drone's current position and the closest point on the roof. Let the number of obstacles in the proximity of the drone be $n_{obstacles}$ and the $d_{obstacles_{i}}, i \in \{1,...,n_{obstacles}\}$ be the distance to the closest point on the boundary of each obstacle from the drone. If any of these distances are less than $2.5~m$, the drone cannot descend at its current position and a new descending position must be computed. The new descending position is computed by iteratively searching for a point along the neighboring regions of the drone's current position. The search is done until a point that is at least $2.5~m$ away from the roof and obstacles is found. The search is done only in the area corresponding to the descent region of the drone. The descent position computed for the sample environment considered is shown in Fig. \ref{fig:seciv_h}.

Once a safe descent point is found, the drone moves to that point while maintaining its current height. Then, the drone begins to descend itself until it reaches a certain height from the ground. In the implementation, the drone was descended to a height of $2~m$ above the ground.

\section{Vision-based Navigation to Door}
\label{sec:navigation}
This section describes the mapping and the routing methods used by the drone to navigate itself to the door of the house after descending. The implementation of this section is used only when the drone needs to deliver the package at a location that is not the descending spot itself. For instance, if the final delivery location specified is the front or the back yard, the drone can descend itself further to land and then, drop the package. Out of the four delivery locations considered in this work, the vision-based navigation to the door is used only when the delivery location is the front door of the house.

\begin{figure*}[t]
\centering
    \begin{subfigure}{0.33\linewidth} % input image
        \includegraphics[width=\linewidth]{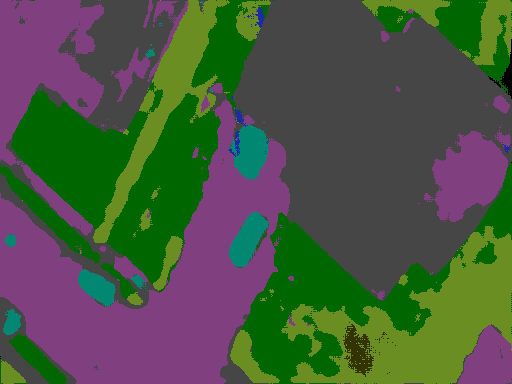}
        \caption{}
        \label{fig:secv_a}
    \end{subfigure}\hfill
    \begin{subfigure}{0.33\linewidth} % input image
        \includegraphics[width=\linewidth]{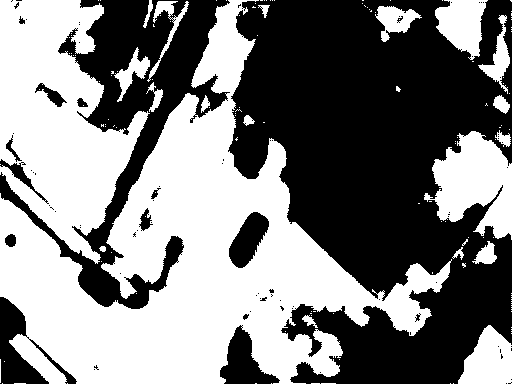}
        \caption{}
        \label{fig:secv_b}
    \end{subfigure}\hfill
    \begin{subfigure}{0.33\linewidth} % input image
        \includegraphics[width=\linewidth]{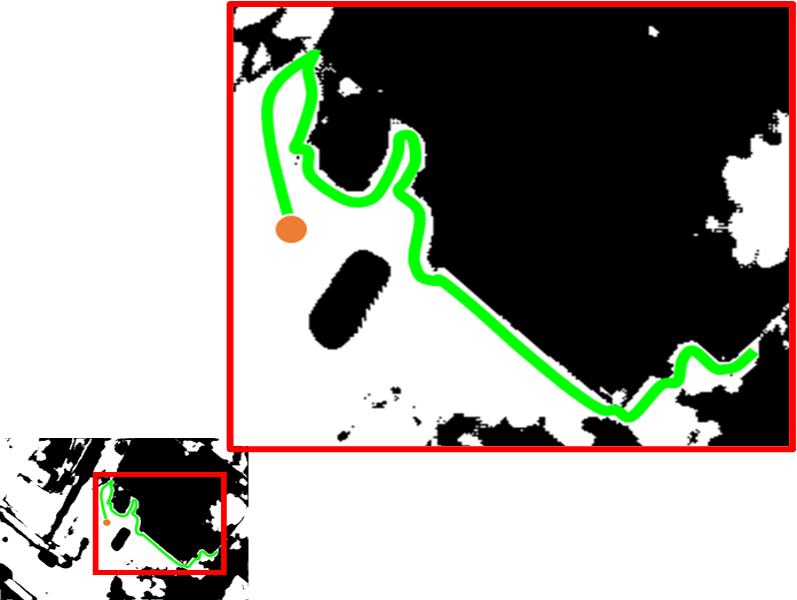}
        \caption{}
        \label{fig:secv_c}
    \end{subfigure}\hfill
\caption{(a) The segmented aerial image captured prior to the drone's descend to the descent spot; (b) The occupancy map generated from the segmented aerial image (black corresponds to the occupied regions and the white corresponds to free empty space); and (c) The path (green line) computed to route the drone along the footprint of the house from the descent spot (orange dot) over the occupancy map.}
\label{fig:cost_map_path}
\end{figure*}

\noindent\textbf{Aerial Occupancy Map.~}
An occupancy map is constructed based on the segmented aerial image captured right before the drone begins descent. The areas corresponding to the roof and other obstacles are considered as occupied regions and the paved area and the grass are considered as open spaces. The height from which the image was captured is used to estimate the resolution of this occupancy grid map. Fig. \ref{fig:secv_a} shows the segmented aerial image prior to the drone's descent to the descent spot and Fig. \ref{fig:secv_b} shows the occupancy map generated from the segmented aerial image (black corresponds to the occupied regions and the white corresponds to free empty space). This occupancy map is used later to compute a path along the boundary of the house's footprint and search the front door of the house. 

% From the occupancy grid map, a path to traverse along the boundary of the house's footprint and open space is computed. Now the drone moves along this path until it finds the door.

\noindent\textbf{Mapping.~}
Once the drone completes the descent, the drone begins to map the environment using the front-facing stereo camera. The environment is mapped as an OctoMap using the point cloud data obtained from the stereo camera. In the implementation, RTAB-Map \cite{labbe2019rtab} was used for estimating the odometry and for building the OctoMap. This OctoMap constructed \footnote{Note: The OctoMap is independent of the occupancy map estimated in the previous step.} is used by the drone later for avoiding obstacles while looking for the front door of the house. 

\noindent\textbf{Door Search.~}
The drone searches for the door visually using the front camera following the descent. In case the door is not visible to the drone right away, the drone performs slight yaw on both sides to extend its viewing range (in the implementation, a yaw range of $[-\pi/4, \pi/4]$ was used). In case, the door of the house is still not detected, the drone navigates along the footprint of the house looking for the door.

The path to route the drone along the footprint of the house is computed using the aerial occupancy map estimated. First, points along the house's footprint are extracted. Then, the path connecting the drone's descent spot and the points along the house's footprint is computed. Fig. \ref{fig:secv_c} shows the path (green line) computed to route the drone along the footprint of the house from the descent spot (orange dot) over the occupancy map obtained from the aerial image. Now, this path is transformed into the frame of the OctoMap and the drone follows this path while visually looking for the door. 

The results of the semantic segmentation may not be accurate at times. This leads to an inaccurate occupancy map. The path computed using this occupancy map can lead the drone to collide with some obstacles. To avoid such collisions, the OctoMap is used to estimate collisions and then, compute a dynamic path avoiding the obstacles. In the implementation, the Robot Operating System (ROS) Navigation Stack was used to compute the dynamic collision-free path. Potentially, the dynamic planner can handle collisions that can occur due to the dynamic changes in the scene too. 

\noindent\textbf{Path Planning to Door.~}
Once the door is found through visual detection, the 3D coordinate of the door in the OctoMap of the environment is estimated. The door's 3D coordinate is estimated by mapping the center of the detected doors bounding box to the point cloud of the environment. To maintain, a safe distance from the door during delivery, the point on the door is offset by a distance of $1~ m$ along the normal direction at that point. This gives a point right in front of the door that is suitable for delivering the package and also resembles the way human couriers deliver a package. In the implementation, Point Cloud Library \cite{Rusu_ICRA2011_PCL} was used for estimating mapping between the image and the point cloud and also for estimating the normal. 

Finally, the drone computes a path to the point in front of the door. Once the drone reaches that point, it descends itself completely so that the package can be delivered. 

\section{Experiments}
\label{sec:expt}
The proposed system was experimented using AirSim\cite{shah2018airsim}, a photo-realistic simulator from Microsoft. Robot Operating System (ROS) was used for the entire implementation. The simulated drone was set up as per the configuration mentioned in Sec. \ref{sec:sys_overview}. An urban neighborhood environment provided as a part of the AirSim simulator was used for the experiments. Houses with various layouts were used in the experiments to show the effectiveness and the robustness of the proposed system. The experiments were repeated by varying the delivery locations and the initial GPS coordinate. The proposed system was compared with a frontier exploration-based method \cite{yamauchi1997frontier} and the time taken to reach the final destination was compared. Frontier exploration is the most commonly used method for exploration and search. The delivery of the package to the front door involves searching for the door after reaching the house. Hence, the frontier exploration-based method was used as a comparison to the proposed method. 

\subsection{Semantic Segmentation}
A semantic segmentation model was trained for obtaining the locations of the various regions from the aerial image during the experiments. %The details about the dataset and the model used are presented below.

\noindent\textbf{Dataset.~}
Semantic Drone Dataset \cite{drone_dataset} was used to train the segmentation model to extract the various features present in the vicinity of the delivery site. The dataset contains aerial images of over 20 houses obtained from an altitude of 5 to 30 $m$ above the ground. The dataset has pixel-accurate annotations of 22 class labels: paved area (road and pavement), dirt, grass, gravel, water, rocks, pool, vegetation, roof, wall, window, door, fence, fence-pole, person, dog, car, bicycles, tree, bald-tree, ar-marker and obstacle (objects other than mentioned earlier) that are critical in deciding a safe spot for the drone to descend its altitude. To train and test the model, we used the public dataset split which contains 400 images. The dataset was split in the ratio of 90:10 for the train and the test images.

\begin{figure}[h]
    \centering
    \includegraphics[width=0.95\columnwidth]{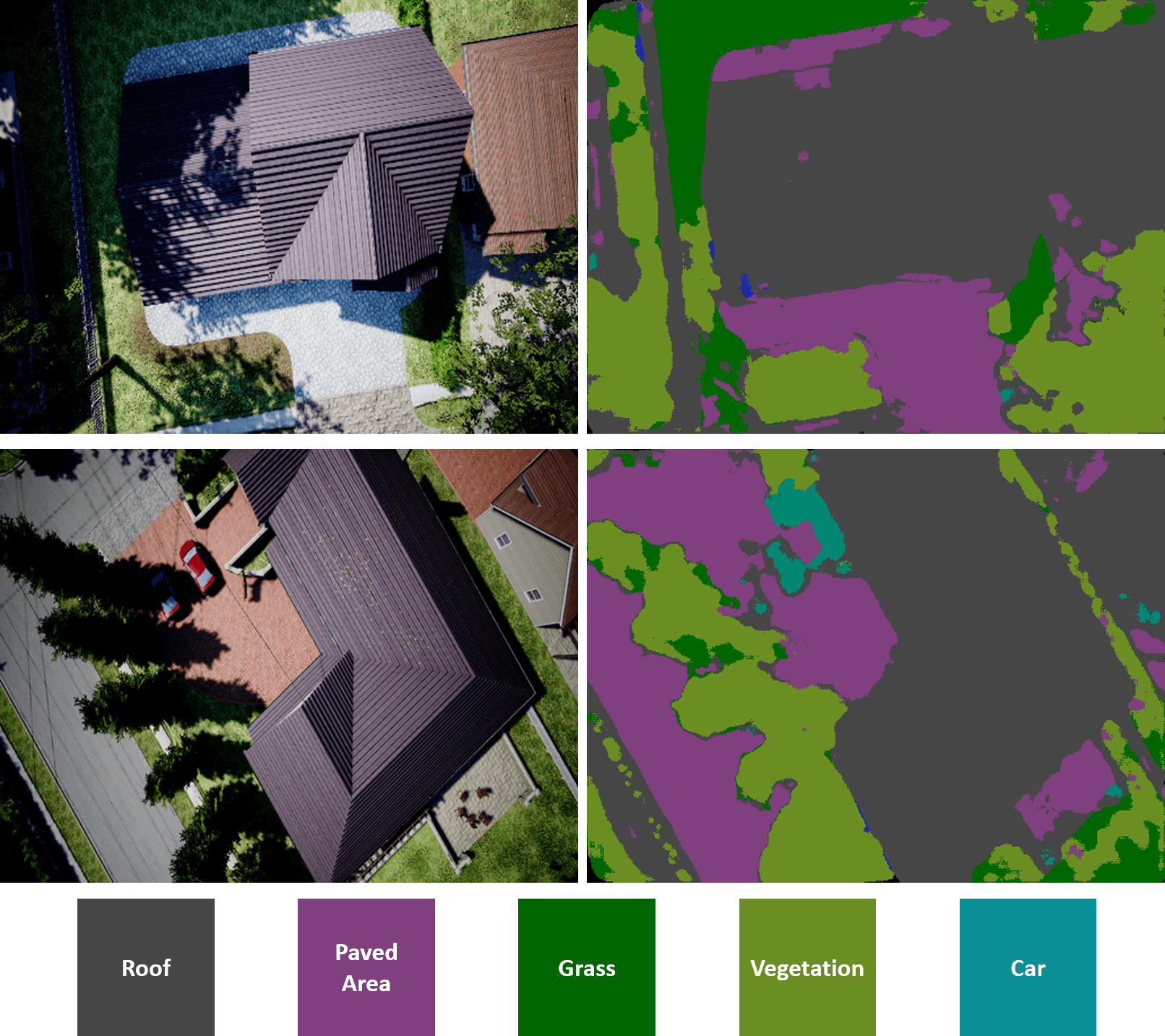}
    \caption{Semantic segmentation results: The left column shows the aerial images captures by the drone in a simulated environment, and the right shows the corresponding segmented images. The labels of the segmentation results are shown at the bottom.}
    \label{fig:sem_seg}
    % \vspace{-7mm}
\end{figure}

\noindent\textbf{Semantic Segmentation using Unet.~}
A semantic segmentation model based on Unet \cite{ronneberger2015u} with VGG-16 \cite{simonyan2014very} as initialization was trained on the dataset. In the implementation, the following classes from the dataset only were considered: roof, paved area, grass, vegetation, fence, car, and tree. The classes were selected based on the objects that are available in the simulated test environment. The model was trained for $50$ epochs using the Adam optimizer. An initial learning rate of $10^{-4}$ was used. Fig. \ref{fig:sem_seg} shows aerial images obtained from the drone (left) in the simulation environment and the segmented images (right) alongside the class labels. Classification accuracy of $83.6\%$ was obtained on the test images using the model trained. The Intersection over Union (IoU) scores for the individual classes for the semantic segmentation model trained on the test images are shown in Table \ref{tab:io_scores}.

\begin{table}[h]
\centering
\caption{Class wise Intersection over Union (IoU) scores for the semantic segmentation model trained for the experiments on the test images.}
\label{tab:io_scores}
\begin{tabular}{|c|c|c|c|c|c|c|}
\hline
\textbf{Class} & \textbf{Roof} & \textbf{Paved-area} & \textbf{Car} & \textbf{Grass} & \textbf{Vegetation} & \textbf{Tree} \\ \hline
\textbf{IoU}   & 0.81          & 0.76                & 0.74         & 0.69           & 0.54                & 0.15          \\ \hline
\end{tabular}
\end{table}

\subsection{Door Detection}
An object detection model was trained to visually detect the door of the house during the experiments. %The details about the dataset and the model used are presented below.

\noindent\textbf{Dataset.~}
DoorDetect dataset \cite{arduengo2019robust} was used to train the door detection model that can identify doors in the image. The dataset contains 1,213 images of various objects such as doors, handles, cabinets, and refrigerator doors collected from various public datasets along with their bounding boxes. In our work, the images of the door (refers to any room door) alone were used to train the detection model.

\noindent\textbf{Object Detection using YOLO.~}
A Convolutional Neural Network (CNN) model based on YOLOv3 \cite{redmon2018yolov3} was used to train the door detection model. The model was trained following an 80:20 split on the dataset. A mean Average Precision (mAP) of $46.3\%$ was obtained on the test dataset. 

\subsection{Quantitative Assessment}
One of the key advantages of the proposed method is that it uses the aerial image captured prior to descending to efficiently find the door in a short time. The time taken by the proposed method to route the drone from the top of the house to the front door of the house was compared with the time taken by a frontier exploration-based method to reach the front door. For the frontier exploration-based method, the drone was set to descend at a random open spot in front of the house and then search for the door using frontier exploration. To prevent the drone from exploring randomly into the roads, the drone was limited in terms of the distance it can explore from the initial location provided. Experiments using the two methods were done on 20 different house setups in the simulated environment. The houses selected for the experiments differed in terms of their size, the kind of features around them and, the visibility of the door. Some of the houses had the door evidently visible from any location in front of the house, whereas some had the door a bit secluded due to the design of the house. Also, the space in front of some houses was cluttered with objects like cars and vegetation and some had an open lawn. The houses with such differences were selected keeping in the mind the differences the drone might face in the real world. The readers are recommended to refer to the supplementary video for images of the houses used in the experiment. 

\begin{figure}[h]
    \centering
    \includegraphics[width=0.8\columnwidth]{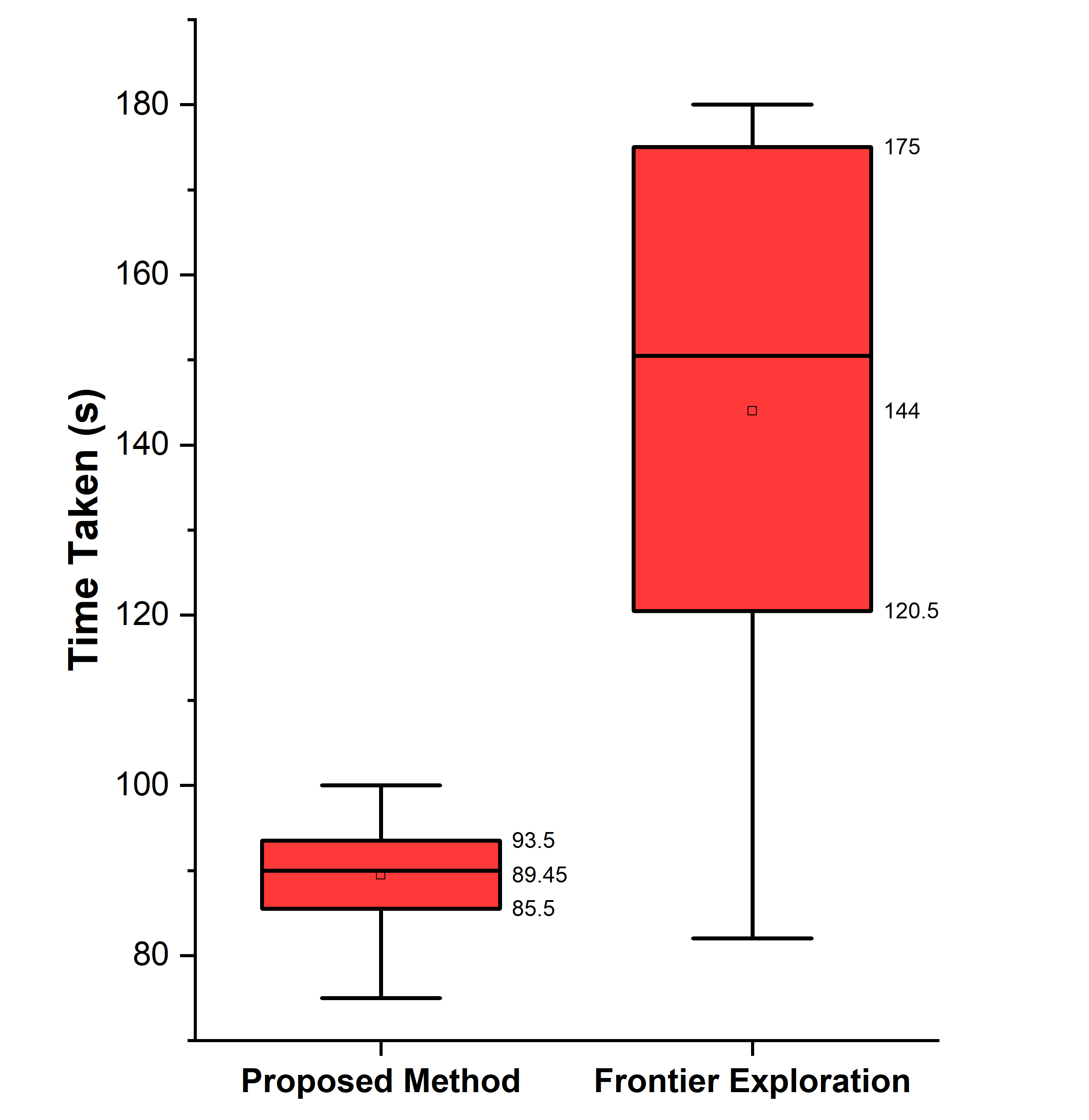}
    \caption{Summary of the time taken by the proposed method and a frontier-based exploration method to route the drone from top of the house to the front door across 20 test houses.}
    \label{fig:comp_plot}
    % \vspace{-7mm}
\end{figure}

Fig. \ref{fig:comp_plot} summarizes the time taken by the drone to reach the door using the two methods across the 20 trials. In four out of the twenty experiments, when there was ample open space around the house, the frontier exploration-based method was exploring regions irrelevant to the footprint of the house and never found the door. For such cases, the maximum exploration time was capped to $180$ seconds. The maximum speed of the drone was set to $0.5$ $m/s$ for all the experiments. 

Across the 20 test houses, the proposed method took an average time of $89.45$ seconds to descend and then reach the door, whereas the frontier exploration-based method took $144$ seconds on average. The proposed system is $161\%$ faster than the frontier on average. The proposed method is faster in terms of the search time since it narrows down the search space by looking for the door primarily along the bounds of the house. While the frontier exploration-based methods do not use any strategy that can lead the drone towards the door of the house quickly. The measures of time taken to reach the final destination had a standard deviation of $6.5$ seconds for the proposed method and $30.5$ seconds for the frontier exploration method. The proposed method has a smaller standard deviation due to its streamlined search method. While in the frontier exploration the drone moves to a new frontier that may or may not lead to the region where the door of the house is. Hence, the door was found in a shorter span during some experiments, whereas it took a long time in some cases since the drone was exploring regions far away from the door. The readers are recommended to refer to the supplementary video for the visuals of the performance using both methods. 

The quantitative comparison was done only for the scenarios where the final destination was the door of the house. The scenarios where the drone needs to reach front yards or back yards were not considered since it only involves descending the drone at a spot and there is not any significant time difference between the two methods. 

\subsection{Empirical Evaluation}
\begin{figure}[!t]
\centering
    \begin{subfigure}{0.975\linewidth} % input image
        \includegraphics[width=\linewidth]{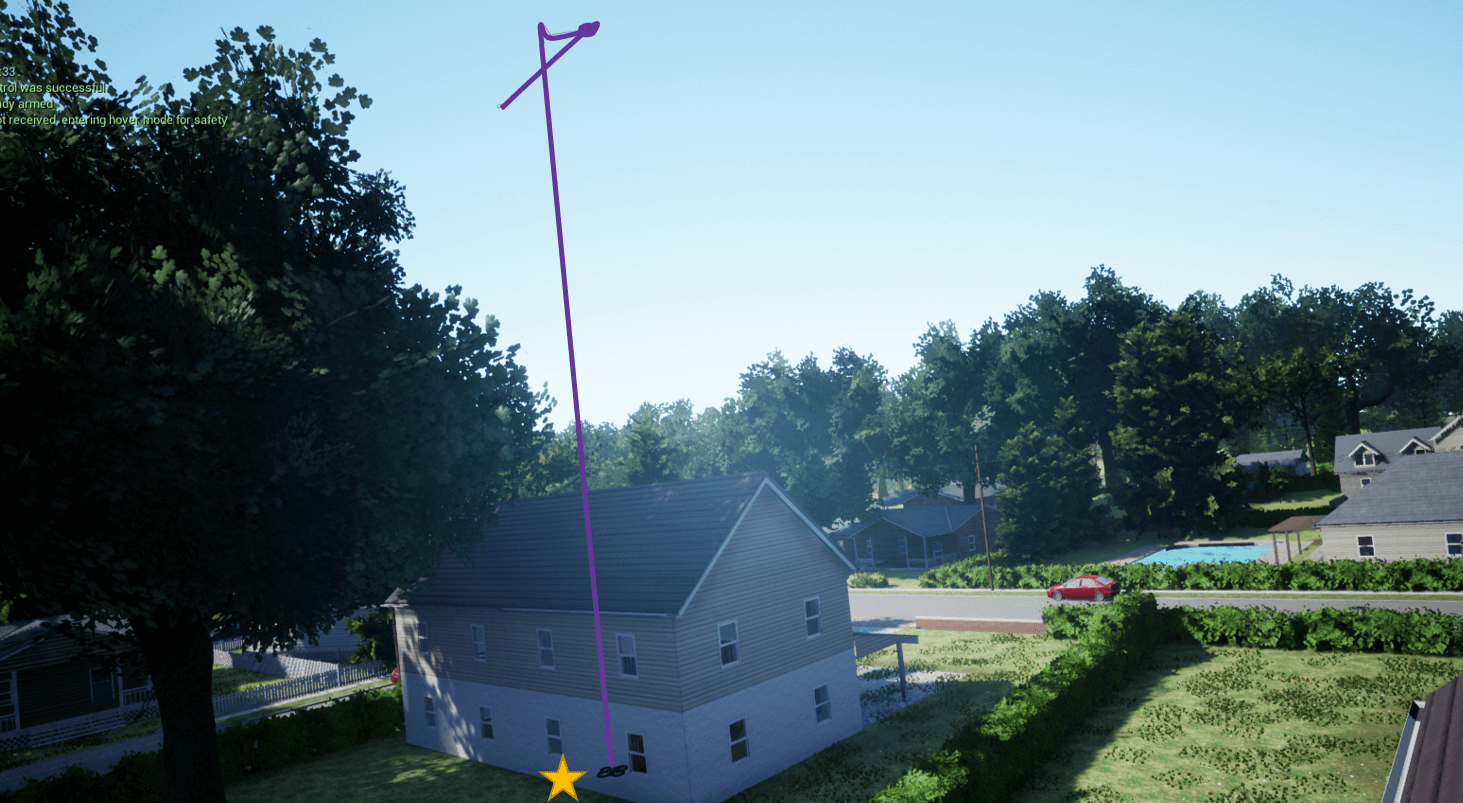}
        \caption{Drone descending to the back yard from a location above the house.}
        \label{fig:back}
    \end{subfigure}\hfill \\
    \vspace{5pt}
    \begin{subfigure}{0.975\linewidth} % input image
        \includegraphics[width=\linewidth]{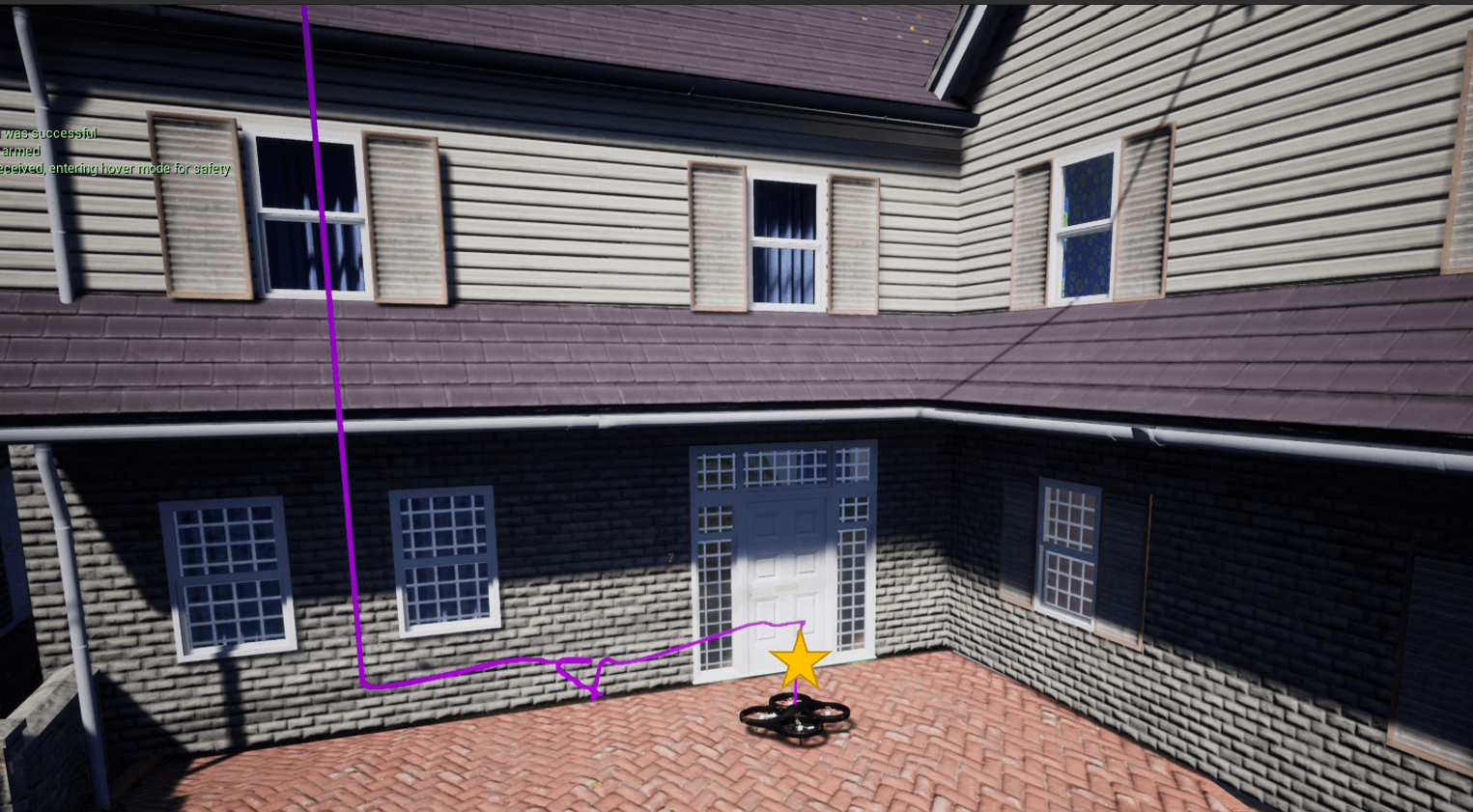}
        \caption{Landed drone along with the trajectory (purple) used for descending and moving right to the door following the descent.}
        \label{fig:house1}
    \end{subfigure}\hfill \\
    \vspace{5pt}
    \begin{subfigure}{0.975\linewidth} % input image
        \includegraphics[width=\linewidth]{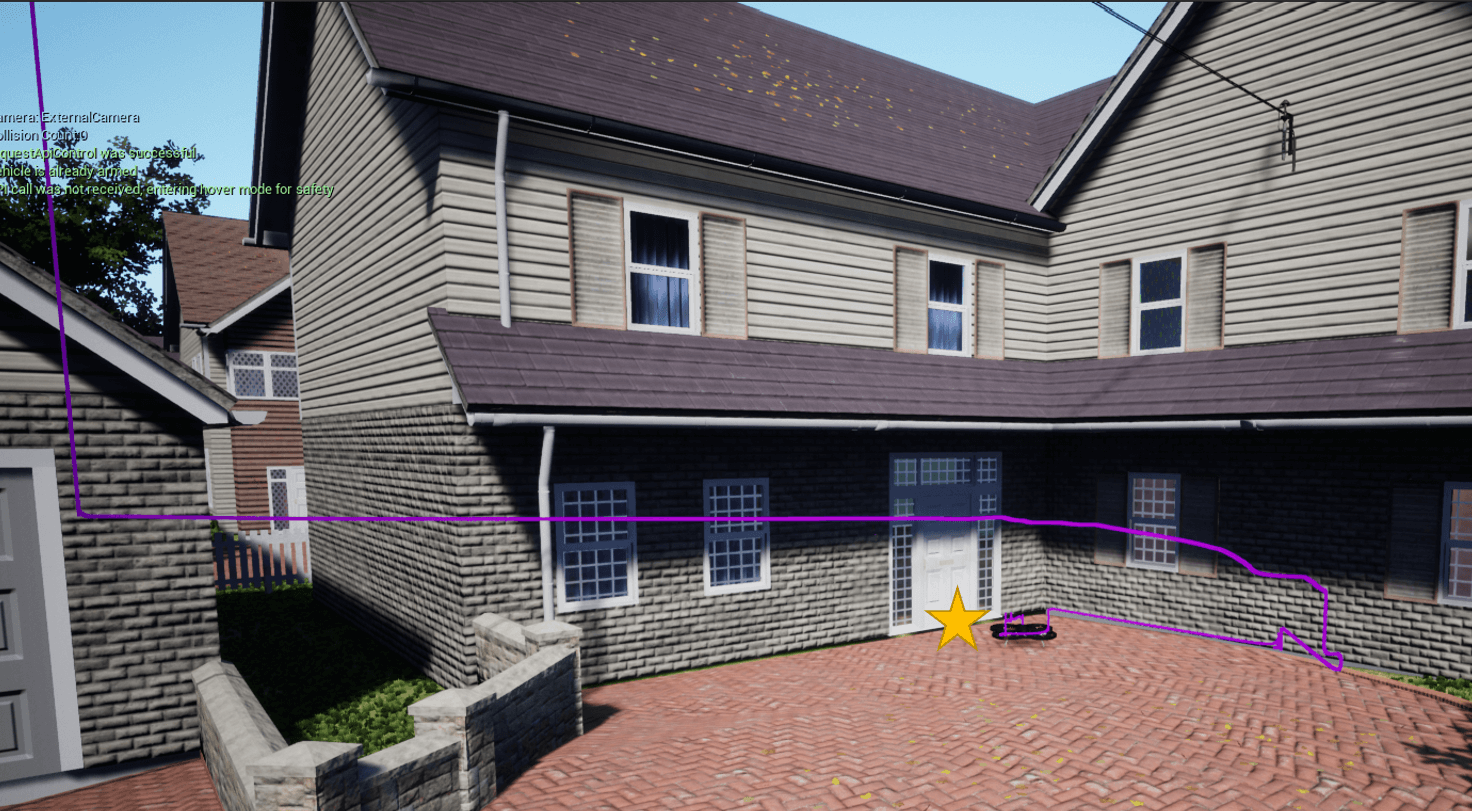}
        \caption{Landed drone along with the trajectory (purple) used for descending and searching for the door along the bounds of the footprint of the house.}
        \label{fig:house2}
    \end{subfigure}\hfill
    \vspace{5pt}
\caption{Drone descending to various regions around the house and delivering the package in front of the door in the simulated environment. The paths traced by the drone in this process are shown in purple. The final delivery spot given to the drone in each of the scenario is marked with $\bigstar$.}
\label{fig:experiment}    
\vspace{-5mm}
\end{figure}

In addition to the quantitative assessment, the performance of the proposed method was evaluated empirically as well with variations in the final delivery location and also the initial starting location on top of the house. For the 20 test houses used in the quantitative assessment, experiments were conducted where the drone needs to deliver the packages to the front paved area, front yard, and back yard depending on the regions that were available in each house. During the experiments, the drone was able to distinguish the back yard from the front yard and find a safe location to descend itself based on the final destination provided the flying height of the drone was high enough to cover both the front and back yards. When the flying height of the drone was close to $30~m$ the aerial image was able to cover both the front and back yard even for larger houses. 
Fig. \ref{fig:back} shows the drone descending into the back yard of a test house along with the trajectory traced by the drone marked in purple. The top part of the trajectory corresponds to the drone aligning itself with the descent location and the following straight line corresponds to the path traced by the drone while descending. Similar results and trajectories were obtained when the final delivery location was specified as the front yard or the front pavement. For brevity, the visual result of descending to the back yard is only presented. 

Multiple initial GPS locations of the drone with the same final destination (front paved area, front yard and, back yard) were also tested. It was observed that changing the initial GPS location of the drone did not contribute to any major changes in the descending behavior of the drone. The drone was always able to deliver the package somewhere close to the footprint of the house. 

Though changing the initial GPS location did not have any significant effect when the final destination was front paved area, front yard or back yard, changes were observed when the final destination was the front door of the house. Irrespective of the initial GPS location, the drone descended itself to the front paved area or the front yard and routed itself to the door. But the way and the path traced by the drone in reaching the front door differed. Fig. \ref{fig:house1} and \ref{fig:house2} show the drone landed in front of the door following the descent  over the paved area. In the experiment shown in Fig. \ref{fig:house1}, the drone descended close to the door, and the door was detected following a minor yaw motion performed by the drone. But, in the experiment shown in \ref{fig:house2} the drone descended close to the garage and was far away from the main door of the house. The door was not detected even following a yaw motion. The drone routed itself along a path following the footprint of the house until it detected the main door. One of the two behaviors of the drone was observed during multiple delivery attempts to the front door of the different houses from various initial GPS coordinates.

\subsection{Discussions}
Overall, our method was successfully able to route the drone to various locations around the house so that it can deliver the package. Following numerous experiments, the method was found to be robust to the numerous variations that the drone might encounter in the real world and also faster compared to using exploration-based methods. The drone was able to find a safe descent location and also route itself to the door. A safe descent location was estimated until there exists a location on the desired descent region that is at a safe distance from the objects and vegetation around. Hence, the system expects is that the descent region of the drone should not too cluttered and that there is not any location safely distanced from the objects around. The system also identified the front door of the house and reached it provided the house follows a typical structure of a single unit house or a townhouse.  With a deeper understanding of the various house architectures, the system can be enhanced to deliver the package to any type of house. Potentially, the system can be even used for delivering packages to apartment-type houses where they can descend on the open spaces around the apartment building. But human intervention might be needed to pick up the package as soon as the drone leaves it. 

As discussed above, the GPS coordinate from which the drone begins the delivery process was found to have a key role in deciding the descent position and also the strategy used by the drone to find the front door of the house. Until the GPS coordinate was close to the proximity of the house, the drone was able to detect the correct roof of the house and the delivery attempts were successful. When the GPS coordinate provided was close to the outer bounds of the house, the roof of the neighboring house was detected as the roof of the recipient's house and subsequently, the whole delivery attempt failed. Hence, it is important that the GPS coordinate provided is not close to the outer bounds of the house, and closer the GPS coordinate is to the center of the house the better it is. 

At times, the system suffered from misclassifications in the semantic segmentation. During the experiments, misclassification frequently occurred over the shadowed regions in the image. In the experiments, it was commonly noticed that the shadowed regions on the roof were often misclassified as paved areas. Similarly, trees were often misclassified  as grass and vice-versa. This led to the poor IoU measure for the trees in Table \ref{tab:io_scores}. These misclassifications can potentially lead to the estimation of descent locations that may not be safe and can cause damage to the package or the objects around. Such issues can be potentially fixed by training the model using a larger dataset and also investigating the development of semantic segmentation models that are more specifically developed for classifying drones' aerial images. During the experiments, the initial height of the drone at the starting GPS coordinate was randomly set between $20$ to $30~m$. In most of the cases, the downward-facing camera of the drone was able to observe the entire house (house and other open spaces around it). However, at times the entire house was not covered in the aerial image and the height of the drone needs to be increased to capture the entire house. But, the increase in the height affected the accuracy of the semantic classifier. Hence, in the development of new aerial image datasets, it is better to capture the images from a wide range of heights. A video demonstration of the experiments is available at: \url{https://youtu.be/WdUqDHH6Emw}.

\section{Conclusion}
\label{sec:conclusion}
In this paper, we present an integrated system that allows drones to autonomously deliver packages at various locations around a recipient's house. The proposed system does not require any direct human interaction during package delivery; the drone autonomously delivers the package at a location specified by the recipient, much as would a human courier. As determined across 20 test houses, the proposed method outperforms the simple frontier exploration-based method by $161\%$ in terms of the time taken to reach the final destination.

Although autonomous delivery drones yet remain a futuristic prospect, our work provides a working proof of concept that addresses the various challenges involved in last-mile package delivery by drones. Future work will focus on conducting real-world experiments with actual human input for the delivery location. Our overall aim is to develop an end-to-end system that allows the recipients to specify delivery instructions in natural language, and drones to deliver packages as per those instructions. In our current implementation, drones are unable to search for house doors when faced with complex house designs in which the door is not clearly visible. We intend to add intelligent search strategies that enable door identification in such scenarios. Furthermore, we intend to improve the accuracy of the aerial image semantic classifier. For one, we plan to develop a larger dataset of aerial images covering different housing environments. In addition, we will develop a novel semantic segmentation neural network capable of segmenting aerial images with higher accuracy. 

%\addtolength{\textheight}{-12cm}   

%\section*{Acknowledgement}
%This work was supported in part by the Purdue Research Foundation (PRF) Graduate Fellowship.

\bibliographystyle{IEEEtran}
\bibliography{references}
\end{document}